\documentclass[conference]{IEEEtran}
\IEEEoverridecommandlockouts
\usepackage{cite}
\usepackage{amsmath,amssymb,amsfonts}
\usepackage{algorithmic}
\usepackage{graphicx}
\usepackage[export]{adjustbox}
\usepackage{textcomp}
\usepackage{xcolor}
\usepackage{amsmath}
\usepackage{amssymb}
\usepackage{cmll}
\usepackage{ upgreek }
\usepackage{multirow}
\usepackage{tabularx}
\usepackage{romannum}
\def\BibTeX{{\rm B\kern-.05em{\sc i\kern-.025em b}\kern-.08em
    T\kern-.1667em\lower.7ex\hbox{E}\kern-.125emX}}

\title{A Lightweight and Interpretable Deepfakes Detection Framework}

\makeatletter
\newcommand{\linebreakand}{%
  \end{@IEEEauthorhalign}
  \hfill\mbox{}\par
  \mbox{}\hfill\begin{@IEEEauthorhalign}
}
\makeatother

\author{\IEEEauthorblockA{Muhammad Umar Farooq}
\IEEEauthorblockA{Department of Software Engineering \\
University of Engineering and Technology\\
Taxila, Pakistan \\
softwareengineerumar@gmail.com}
\and
\IEEEauthorblockA{Ali Javed}
\IEEEauthorblockA{Department of Computer Science \\
University of Engineering and Technology\\
Taxila, Pakistan \\
ali.javed@uettaxila.edu.pk}
\and
\IEEEauthorblockA{Khalid Mahmood Malik}
\IEEEauthorblockA{Department of CS and Engineering \\
Oakland University\\
Rochester, MI, USA \\
mahmood@oakland.edu}
\linebreakand 
\IEEEauthorblockA{Muhammad Anas Raza}
\IEEEauthorblockA{Department of Mechanical Engineering \\
University of Engineering and Technology\\
Taxila, Pakistan \\
memanasraza@gmail.com}
}
\begin{document}

\maketitle
\begin{abstract}
The recent realistic creation and dissemination of so-called deepfakes poses a serious threat to social life, civil rest, and law. Celebrity defaming, election manipulation, and deepfakes as evidence in court of law are few potential consequences of deepfakes. The availability of open source trained models based on modern frameworks such as PyTorch or TensorFlow, video manipulations Apps such as FaceApp and REFACE, and economical computing infrastructure has easen the creation of deepfakes. Most of the existing detectors focus on detecting either face-swap, lip-sync, or puppet master deepfakes, but a unified framework to detect all three types of deepfakes is hardly explored. This paper presents a unified framework that exploits the power of proposed feature fusion of hybrid facial landmarks and our novel heart rate features for detection of all types of deepfakes. We propose novel heart rate features and fused them with the facial landmark features to better extract the facial artifacts of fake videos and natural variations available in the original videos. We used these features to train a light-weight XGBoost to classify between the deepfake and bonafide videos.  We evaluated the performance of our framework on the world leaders  dataset (WLDR) that contains all types of deepfakes. Experimental results illustrate that the proposed framework offers superior detection performance over the comparative deepfakes detection methods. Performance comparison of our framework against the LSTM-FCN, a candidate of deep learning model, shows that proposed model achieves similar results, however, it is more interpretable. 
\end{abstract}

\begin{IEEEkeywords}
Deepfakes, Multimedia Forensics, Random Forest Ensembles, Tree boosting, XGBoost, Faceswap, Lip sync, Puppet Master.
\end{IEEEkeywords}

\section{Introduction}
Recent advancements in deep learning (DL) have impacted the way we solve complex technical problems in computer vision (CV) and robotics. With the widespread availability of video synthesis repositories and video manipulations Apps such as FaceApp \cite{b1} and REFACE \cite{b2}, video manipulation has become easy, even for a layman. Video synthesis is beneficial in some ways like avatar creation, animated video content creation, etc. Sometimes videos are synthesized just for the sake of fun, like a recent realistic Tiktok video of Tom Cruise \cite{b3}. However, the case is not always that simple. Depending on the time and context, deepfakes pose a serious threat to society. With deepfakes, celebrities are defamed, and election campaigns could be manipulated. DL based video synthesis tools use generative adversarial networks (GAN) under the hood. The adaptive nature of GAN made it difficult to develop a robust detection solution. Whenever a deepfakes detection model is developed, we witness some variant of a GAN based generation model to exploit the newly developed detection model by manipulating its cues. Thus, deepfakes creation and detection is a constant battle between the ethical and unethical machine learning (ML) experts.

Deepfakes detection got much attention in the last decade after realistic fake videos of politicians and celebrities got viral via social media platforms. Current deepfake videos are categorized as face-swap, lip-sync, and puppet master \cite{b4}. In face-swap deepfakes, face of a target person is added at the place of a source person in the original video to create a fake video of the target person. In lip-sync deepfakes, lips of a person are synced for an audio to reflect that person is speaking the text in that audio. In puppet-master, the face of the target person is placed in the original video but facial expressions of the source person are retained on the target face to make the fake more realistic. Most of the existing detection solutions target specific types of deepfakes, however, generic solutions capable of countering all types of deepfakes are less explored. For example, Agarwal et al. \cite{b5} proposed a detection technique for lip-sync deepfakes. This technique exploited the inconsistencies between the viseme (mouth shape) and a phoneme (spoken word). This work applied manual and CNN based techniques to compute the mapping of viseme to phoneme. This model is good for a specific set of seen data. However, model performance can degrade on unseen data for different patterns of viseme to phoneme mapping, with the change of speaking accent or even non-alignment of audio-to-video. 

Most of the existing systems are unable to perform well on all three types of deepfakes. Moreover, deepfakes detection models based on the traditional classifiers like SVM, works only where data is linearly separable. CNN based models are computationally more complex and are black-box in terms of prediction. Therefore, this paper addresses the following research questions:
\begin{enumerate}
\item Is it possible to improve the detection accuracy of deepfakes using hybrid landmark and heart-rate features on a diverse dataset containing all three types of deepfakes?
\item Is it possible to create a generalized detection model based on proposed hybrid landmark and heart-rate features and ensemble learning?
\item Is it possible to achieve the same accuracy as deep learning models but improve the interpretability by using an ensemble of supervised learning?
\end{enumerate}

Existing deepfake detection techniques are broadly categorized as handcrafted features \cite{b6,b7,b8,b9} based or DL based \cite{b10,b11,b12,b13,b14}. For example, Yang et al. \cite{b9} used 68-D facial landmark features to train an SVM classifier for detection. This work achieved good performance on good quality videos of UADFV \cite{b9} and DARPA MediFor \cite{b15} datasets but was unable to perform well on low quality videos. Moreover, the evaluation of this work did not consider all types of deepfakes. Matern et al. \cite{b6} used 16-D texture based eyes and teeth features for the exploitation of the visual artifacts to detect video forgeries like face-swap and Face2Face. Most important aspect of this work was to detect the difference in eye color of a POI for detection of face-swap deepfakes detection by exploiting the missing details like reflection in eye color. Additionally, this work uses face border and nose tip features along with eye color features for Face2Face deepfakes detection. This technique \cite{b6} has a limitation of working only for faces with clear teeth and open eyes.  Lastly, the evaluation of this work was only performed on FaceForensics++ \cite{b10} dataset. Li et al. \cite{b7} used the targeted affine warping artifacts introduced during deepfakes generation. Targeting the specific artifacts reduced the training overhead and improved the efficiency. However, these specific artifacts selection can compromise the robustness of this technique by making it difficult to detect a deepfake with slightly new transformation artifacts. Agarwal et al. \cite{b8} used an open source toolkit OpenFace2 \cite{b16} for facial landmark features extraction. Some features were derived based on extracted landmark features. These derived features were then used along with action unit (AU) features to train a binary SVM for deepfakes detection. This technique was proposed for five POIs where all POIs were linearly separable in a t-SNE plot. However, for an increased number of POIs in the updated dataset \cite{b17}, performance of this technique was significantly degraded. In their extended work, Agarwal et al. [18] proposed a framework based on spatial and temporal artifacts in deepfakes. This framework is based on some threshold based rules to classify a video as real or fake. This rule-based approach would work on selected datasets, however, performance of this hard coded threshold oriented approach is expected to degrade on unseen data.  In \cite{b18}, authors proposed a new framework, ‘FakeCatcher’, which uses biological signals from three face regions in the real videos to detect the fake videos. FakeCatcher applied many transformations on biological features like autocorrelation, power spectral density, wavelet transform, etc. Authenticity decision is based on the aggregated probabilities of two probabilistic classifiers (SVM and CNN). Performance was evaluated on their own customized dataset, however, it is not evaluated on all three types of deepfakes.

Besides the handcrafted features-based methods, deep learning-based methods are also being employed for deepfakes detection. Guera et al. \cite{b10} applied a DL based technique to detect the deepfakes. This technique applied a CNN to extract features followed by a long-short term memory (LSTM) to learn those features. Important contribution of this work was the exploitation of temporal inconsistencies among deepfakes for classification. However, this approach is unable to identify all three types of deepfakes. Afchar et al. \cite{b11} designed a neural network (MesoNet) to detect deepfakes and Face2Face video forgeries. This work designed an end-to-end architecture with convolutional and pooling layers for feature extraction followed by dense layers for classification. These methods \cite{b10,b11} were evaluated on videos collected from random websites rather than a standard dataset that doubted the robustness of these approaches for a large-scale and diverse standard dataset. Nguyen et al. \cite{b12} designed a capsule network to expose multiple types of tampering in images and videos. This framework aimed at detection of face swapping, facial re-enhancements and computer generated images. This framework used dynamic routing and expectation-maximization algorithms for performance improvement. The Capsule network employed the VGG-19 for latent face features extraction and used them for classification of original and bonafide videos. Framework is good at detecting face-swap forgeries in FaceForensics dataset, however, not evaluated on lip-sync and pupper-master deepfakes and complex in terms of computations. Sabir et al. \cite{b13} proposed a method based on DL to feed cropped and aligned faces to a CNN (ResNet and DenseNet) for feature extraction followed by an RNN for classification. Most important aspect of this work was to use features from multiple levels of CNN to incorporate mesoscopic level features extraction. This work \cite{b13} only used FaceForensics++[11] dataset for evaluation and didn’t consider lip-sync and puppet-master deepfakes. Yu et al. \cite{b14} used a CNN to capture the fingerprints of GAN generated images to perform the classification of synthetic and real images. This technique targeted fake images generated with four GAN variants ProGAN, SNGAN, CramerGAN, MMDGAN, but might not be able to detect fake images generated with a new GAN variant. In \cite{b19}, authors used an ensemble of four CNNs to achieve good results on DFDC. An attention mechanism was added to EfficientNetB4 to get the insights of the training process. EfficientNetB4 and EfficientNetB4Att were trained as end-to-end training, whereas, EfficientNetB4ST and EfficientNetB4AttST were trained in Siamese training settings. Important aspect of this method was the data augmentation (i.e. down sampling, hue saturation, JPEG compression, etc.) during training and validation for model robustness. Moreover, this technique performs well on large face-swap dataset DFDC but not evaluated on all three types of deepfakes and is computationally complex. In \cite{b20}, authors used EfficientNet (a CNN) and gated recurrent unit (GRU) (an RNN) to exploit spatiotemporal features in the video frames to detect deepfake videos. This work included data augmentation on real videos during training to balance classes as DFDC is highly class imbalanced.  Moreover, this architecture performs well on large face-swap dataset DFDC but is not evaluated on all three types of deepfakes and is complex in terms of computations.

Most methods based on handcrafted features \cite{b6,b7,b8,b9} fail to generalize well on different types of deepfakes like lip-sync and puppet-master. CNN based techniques \cite{b10,b11,b12,b13,b14} are computationally complex and black-box in terms of generating the output. Moreover, these methods exploit some GAN specific artifacts produced during generation. So, they might fail to detect deepfakes, generated with a new GAN architecture.

To address the above mentioned problems and limitations of existing works, this paper proposes a lightweight model based on feature fusion of facial landmarks and heart rate features. For landmark features, we analyzed the impact of each landmark features category before final features selection. We analyzed the impact of different combinations of features categories. We  started with two most effective features categories and then added one category in the feature-set at a time in the decreasing order of effectiveness. We disregarded the concept of the POIs being linearly separable, because that concept becomes invalid with a higher number of POIs. We used the XGBoost \cite{b21} for classification purposes. XGBoost uses Bagging in Random Forest for variance related errors and gradient boosting algorithm for bias related errors. XGBoost successfully addresses the data classification problem where data points are not linearly separable. 

The main contributions of this paper are as follows:
\begin{itemize}
\item We propose a lightweight and interpretable deepfakes detection framework capable of accurately detecting all types of deepfakes namely, faceswap, puppet-master and lipsync.
\item We propose novel heart rate features and fused them with a robust set of selected facial landmark features for deepfakes detection.
\item We highlight that an XGBoost based solution is lightweight as compared to CNN based solutions and better generalize as compared to other conventional classifiers  like SVM, KNN, etc.
\end{itemize}
Rest of the paper is structured as follows. Section 2 presents the details of feature engineering and model development. In Section 3, we present the details of performance evaluation and comparative analysis w.r.t to state of the art methods. Finally, we conclude our paper in Section 4.

\section{Methodology}
This section provides an overview of the proposed framework. As shown in the Figure \Romannum{1}, the input video is processed to extract 850-D facial landmarks and 63-D heart rate features. XGBoost classifier is used for classification. Classifier is trained on each sub-category of landmark and heart rate features. Finally, we reduce the dimensions of our features to select the most reliable features among all to make the final features-set. XGBoost classifier is trained on the final features-set to classify the video as fake or bonafide. The process flow of the proposed solution is shown in Figure \Romannum{1}.

\subsection{Features Extraction}

Effective features extraction is crucial for any classification task. For this purpose, we proposed a fused features-set consisting of our novel heart rate features and the facial landmark features. We extracted facial Landmark features using the OpenFace2 [21] toolkit. For heart rate features, we selected seven regions of interest as shown in Figure \Romannum{1}. Seven ROIs are right cheek (RC), left cheek (LC), chin (C), forehead (F), outer right edge (OR), outer left edge (OL), and center (C). We calculated RGB values of all ROIs and then applied some transformations to create heart rate features. Details of transformations are as follows:

\begin{equation}
HR_s=\{Z_R,Z_G,Z_B\} \label{eq}
\end{equation}

\begin{equation}
HR_r=\{Z_R/Z_G,Z_R/Z_G,Z_G/Z_B\} \label{eq}
\end{equation}
Where $HR_s \in {RC,LC,C,F,OR,OL,CE} \with R \leftrightarrow red,G \leftrightarrow green,B \leftrightarrow blue.$

\begin{equation}
HR_r=HR_s  {\displaystyle \cup }  HR_r \label{eq}
\end{equation}
Where $HR_s$ represent the simple heart rate features at ROIs and $HR_r$ is the ratios of heart rate features. Union of these HR features generate our heart rate features.

\renewcommand{\figurename}{Figure }
\renewcommand{\thefigure}{\Roman{figure}}
\begin{figure*}[t]
    \centering
    \includegraphics[width=0.95\textwidth]{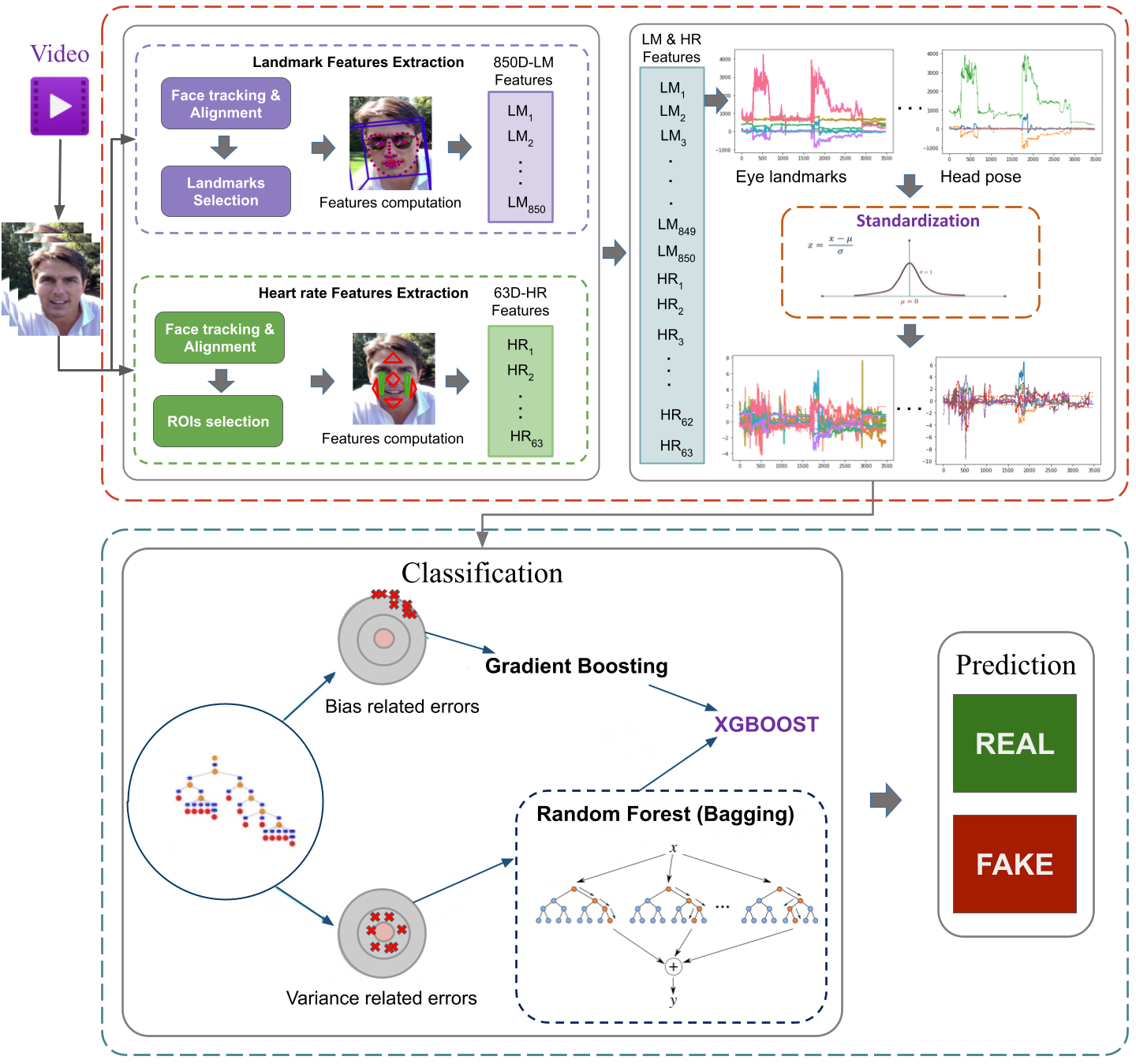}
    \caption{Architecture of the Proposed Framework}
\end{figure*}


\subsection{Features Standardization $\&$ Segmentation}\label{AA}
Both the landmark and our proposed heart rate features are on different scales. To fuse the features, we standardized features by partially learning the distribution of features during data loading. We apply standardization as shown in Eq. (4), based on learned distribution over all features.
\begin{equation}
z=\frac{x-\upmu}{\sigma} \label{eq}
\end{equation}
Where $\upmu$ is mean and $\sigma$ is standard deviation of a feature column.

Our solution works at both the frame and segment level. For segment level operation, we created segments with a length of 30 frames with an overlapping of 10 frames. In our case, the video frame rate is 30 frames per second. 

\subsection{Classification}\label{AA}
For the classification task, we need a classifier that should be lightweight and can generalize easily to the new datasets. Classification process should be interpretable so we can follow a directed path for further improvements. To incorporate those requirements, we employed the extreme gradient boosting (XGBoost) \cite{b21}, an approach for gradient boosted decision trees. XGBoost is an algorithm in the class of gradient boosting machines. In boosting algorithms, many weak learners are ensembled sequentially to create a strong learner having low variance and high accuracy. In boosting, learning of the next predictor is improved to avoid repeating the error caused by any previous predictor. In Random forest, a model with deeper trees gives good performance but in XGBoost, shallow trees perform better because of boosting. There are two boosting approaches, Adaptive Boosting and Gradient Boosting. Adaptive boosting puts more weight on misclassified data samples. While gradient boosting identifies misclassified samples as gradients using the Gradient Descent to iteratively optimize the loss. XGBoost employs Gradient boosting. Using XGBoost will be highly effective for large datasets as it is highly scalable and computationally efficient. We can use the power of GPU as XGBoost can perform out-of-core computations. Objective function of XGBoost is based on training loss and a regularization function as shown in Eqs. (5) $\with$ (6). Training loss helps in stage wise bagging of trees in the random forest to decrease the variance error. Regularization function helps to reduce the bias related errors using boosting. 

\begin{equation}
    \mathcal{O} = \sum_{i=1}^{n} l(y_i,\hat y_i^t) + \sum_{i=1}^{t} \Omega(f_i)
\end{equation}
Where $t$ is the total number of trees and $y_i$ is actual value and  $\hat y_i^t$ is the prediction at time $t$. $n$ is the total number of training samples.

\begin{equation}
    \Omega(f_i) = \gamma T + \frac{1}{2}\uplambda \sum_{j=1}^{T} \omega _j^2
\end{equation}
Where $\gamma$ is the min reduction in loss, required for a new split on leaf node and $\uplambda$ is the $l2$ regularization term on leaf weights and helps ovoid overfitting. 

\section{Experiments and Results}
\subsection{Dataset}\label{AA}
We evaluated our method on the world leaders dataset (WLDR) \cite{b8}. WLDR is the only dataset with all three types of deepfakes. WLDR comprises real and fake videos of ten U.S. politicians, and real videos of comedy impersonators of those political figures. The WLDR dataset has all three types of deepfakes i.e., face-swap, puppet-master and lip-sync. WLDR has lip-sync deepfakes for only one POI i.e., Obama. Face-swap videos of WLDR are created by replacing the face of the impersonator with the face of the corresponding politician. The WLDR dataset has 1753 real and 93 fake videos. Other datasets like DFDC, FF++ and DFD have more fake videos as compared to real videos. WLDR has more real videos (95$\%$) than the fake videos (5$\%$) which is good, as for better detection, we have to learn the patterns in the real videos rather than fake videos as fake videos are constantly changing with the evolution of GANs. Still it is not large enough to generalize a model to perform well in the wild deepfakes. We used area under the curve (AUC) as an evaluation metric for model evaluation. The reason behind using AUC is that almost all the available datasets are highly class imbalanced. AUC gives a fair performance score for imbalanced classes as compared to Accuracy. 

\subsection{Performance Evaluation of Proposed Framework}\label{AA}
The objective of this experiment is to evaluate the performance of the proposed framework on a diverse dataset WLDR, having all three types of deepfakes. For this purpose, we fed the proposed features of selected landmarks and heart rate features to train the XGBoost based random forest ensemble to perform the classification of bonafide and deepfakes. Heart rate features and sub-categories of landmark features are on different scales. We standardized features before feeding to the classifier. For standardization, we calculated mean and standard deviation of the whole training set during data preparation. We scaled train, test and validation sets to make sure the mean of rescaled data is zero and standard deviation is one. We evaluated our model on frame and segment level. In WLDR, the frame rate of videos is 30 frames per second. For segment level evaluation, we created 30 frames length segments with an overlapping of 10 frames. Our model is robust to both frame and segment level detection.

We evaluated our model on each of six categories of facial landmark and heart rate features. List of features effective to the detection task in descending order is 2D landmark, 3D landmark, eye landmark, headpose, heart rate, shape and action unit features. Table 1 presents the results of individual feature types. We conducted an evaluation on different combinations of features in the descending order of their effectiveness. Table 2 presents the results of a combination of features categories. We observed from Table 2 that 2D and 3D lankmark features are most effective giving an AUC of 0.9311. We also observed that eye landmark and headpose features are effective thereby increasing AUC from 0.9311 to 0.9326,  when combined with 2D and 3D landmarks. Additionally, we observed that combining heart rate features with selected landmark features is very effective and increases the AUC from 0.9326 to 0.9505. Based on our observation, we didn’t include shape features in the final features-set due to slight improvement in AUC from 0.9505 to 0.9510 when shape features are also included in the fused features-set. Our final features-set includes eye landmarks, headpose, 2D $\with$ 3D landmarks and heart rate features. As per our hypothesis, combinations of features that are individually effective also perform better. Finally, we selected five out of seven features categories for our model. We evaluated our model on a wide range of parameters. More specifically, we set the learning rate to 0.01, number of trees to 1500, Max depth tree to 8.
\begin{table*}[t]
\caption{Segment and frame level AUC on individual features categories}
\begin{center}
\begin{adjustbox}{width=1\textwidth}
\begin{tabular}{|c|c|c|c|c|c|c|c|}
\hline 
\textbf{Features Used} & \textbf{Eye landmark}& \textbf{Head pose}& \textbf{\textit{2D landmark}}& 
\textbf{\textit{3D landmark}}& \textbf{\textit{Shape}}& 
\textbf{\textit{Action Unit}}& \textbf{\textit{Heart Rate}} \\ 
\hline
(AUC): Seg level & 0.8851 & 0.8023 & 0.8982 & 0.8978 & 0.7644 & 0.5027 & 0.7956 \\
\hline
(AUC): Frame level & 0.8659 & 0.7774 & 0.8903 & 0.8856 & 0.7357 & 0.5017 & 0.7866 \\
\hline
\end{tabular}
\end{adjustbox}
\label{tab1}
\end{center}
\end{table*}
\begin{table*}[t]
\caption{Segment and frame level AUC on combination features categories}
\begin{center}
\begin{adjustbox}{width=1\textwidth}
\begin{tabular}{|c|c|c|c|c|c|}
\hline 
\textbf{Features Used} & \textbf{2D lmk,3D lmk}& \textbf{Eye lmk,2D lmk,3D lmk}& \textbf{\textit{Eye lmk,Headpose,2D lmk,3D lmk}}& 
\textbf{\textit{Eye lmk,Head pose,2D lmk,3D lmk,HR}}& \textbf{\textit{Eye lmk,Head pose,2D lmk,3D lmk,Shape,HR}} \\ 
\hline
(AUC): Seg level & 0.9297 & 0.9311 & 0.9326 & 0.9505 & 0.9510 \\
\hline
(AUC): Frame level & 0.9158 & 0.9059 & 0.9068 & 0.9425 & 0.9285 \\
\hline
\end{tabular}
\end{adjustbox}
\label{tab1}
\end{center}
\end{table*}
\begin{table}[htbp]
\caption{Comparison of XGBoost with [8],[18] and LSTM-FCN [23]}
\begin{center}
\begin{tabular}{|c|c|c|c|}
\hline 
\textbf{Model Name} & \textbf{WLDR}& \textbf{Evaluation Levels} \\ 
\hline
Protecting World Leaders [8] & 0.93 & Frame and segment level \\
\hline
LSTM-FCN & 0.95 & segment level \\
\hline
XGBoost (proposed) & 0.95 & Frame and segment level \\
\hline
Appearance and Behavior [18] & 0.99 & Video level \\
\hline
\end{tabular}
\label{tab1}
\end{center}
\end{table}

\subsection{Performance Comparison of the Proposed and Existing Methods}\label{AA}
This experiment is designed to measure the performance of our framework against existing state-of-the-art deepfakes detection methods. For this, we compared the performance of the proposed framework against the \cite{b8} and \cite{b17}. Table 3 presents the results of comparison of proposed framework against existing models. Our model outperforms [8] that is based on action unit features and derived features capturing mouth movements but our model performance is lower than their extended work [18]. We also compared our model with a deep learning (DL) classifier, LSTM-FCN \cite{b22}. Agarwal et al. \cite{b8} technique works on the assumption of linear separability of bonafide and deepfake videos in a t-SNE plot based on selected features. But this technique failed to generalize on all types of deefakes. In their extended work, Agarwal et al. \cite{b17} evaluated their method on 10-second video clips rather than frames and segments of small length. Although, this model performs better and generalizes well on all existing datasets of face-swap. However, in this work only face-swap deepfakes are considered and lip-sync and puppet-master deepfakes are not addressed. Moreover, performance of this method \cite{b17} is expected to drop if evaluated on frame and segment level due to its threshold based approach. We observed from the results (Table 3) that a DL based model, LSTM-FCN can achieve comparable results as we achieved with XGBoost based Random Forest ensembles. However, compared to LSTM-FCN our proposed framework is light weight and interpretable rather than a black-box oriented model of a DL classifier.

\section{Conclusion and Future Work}
This work has presented a unified method based on fusion of our novel heart features and facial landmarks for detecting all three types of deepfakes. Unlike many existing methods, our method is light weight, interpretable and effective at the same time. Moreover, compared to existing light weight techniques, our method is more robust and interpretable. We highlighted that an XGBoost based framework is lightweight over the CNN based solutions and generalizes better as compared to other conventional classifiers. For this purpose, we compared our proposed method with a time-series DL classification model, LSTM-FCN. However, proposed framework follows a signature based approach and thus may not be very effective against deepfakes developed in future. Proposed method also need to be enhanced for optimized cross corpus evaluation. For our future work, we’ll perform cross-dataset evaluation, experimenting on the datasets that have multiple forgeries per sample.

\section{Acknowledgments}
This work was supported by grant of Punjab HEC of Pakistan via Award No. (PHEC/ARA/PIRCA/20527/21).

\end{document}